\documentclass[10pt,twocolumn,letterpaper]{article}

\usepackage{iccv}
\usepackage{times}
\usepackage{epsfig}
\usepackage{graphicx}
\usepackage{amsmath}
\usepackage{amssymb}
\usepackage{bm}
\usepackage{caption}
\usepackage{multirow}

\usepackage[accsupp]{axessibility}

\usepackage[breaklinks=true,bookmarks=false]{hyperref}

\iccvfinalcopy 

\def\iccvPaperID{7035} 

\newcommand\blfootnote[1]{%
  \begingroup
  \renewcommand\thefootnote{}\footnote{#1}%
  \addtocounter{footnote}{-1}%
  \endgroup
}

\ificcvfinal\fi

\begin{document}

\title{When do GANs replicate? On the choice of dataset size}

\author{Qianli Feng$^{1,2}$~~~~ Chenqi Guo$^{1}$~~~~ Fabian Benitez-Quiroz$^{1}$~~~~ Aleix Martinez$^{1,2}$\\
    $^1$The Ohio State University~~~~$^2$Amazon~~~~\\
    {\tt\small \{feng.559, guo.1648, benitez-quiroz.1, martinez.158\}@osu.edu}
}


\twocolumn[{%
\renewcommand\twocolumn[1][]{#1}%
\maketitle

\ificcvfinal\thispagestyle{empty}\fi

}]


\begin{abstract}
   Do GANs replicate training images? Previous studies have shown that GANs do not seem to replicate training data without significant change in the training procedure. This leads to a series of research on the exact condition needed for GANs to overfit to the training data. Although a number of factors has been theoretically or empirically identified, the effect of dataset size and complexity on GANs replication is still unknown. With empirical evidence from BigGAN and StyleGAN2, on datasets CelebA, Flower and LSUN-bedroom, we show that dataset size and its complexity play an important role in GANs replication and perceptual quality of the generated images. We further quantify this relationship, discovering that replication percentage decays exponentially with respect to dataset size and complexity, with a shared decaying factor across GAN-dataset combinations. Meanwhile, the perceptual image quality follows a U-shape trend w.r.t dataset size. This finding leads to a practical tool for one-shot estimation on minimal dataset size to prevent GAN replication which can be used to guide datasets construction and selection.\blfootnote{Qianli Feng and Chenqi Guo contribute equally to the paper. Code is available at \url{https://github.com/chenqiguo/GAN_replication}. Authors were supported by NIH grant R01DC014498, R01EY020834 and HFSP grant RGP0036/2016.}
\end{abstract}


\section{Introduction}\label{sec:intro}

\begin{figure}[h!]
\begin{center}
\includegraphics[width=1.0\linewidth]{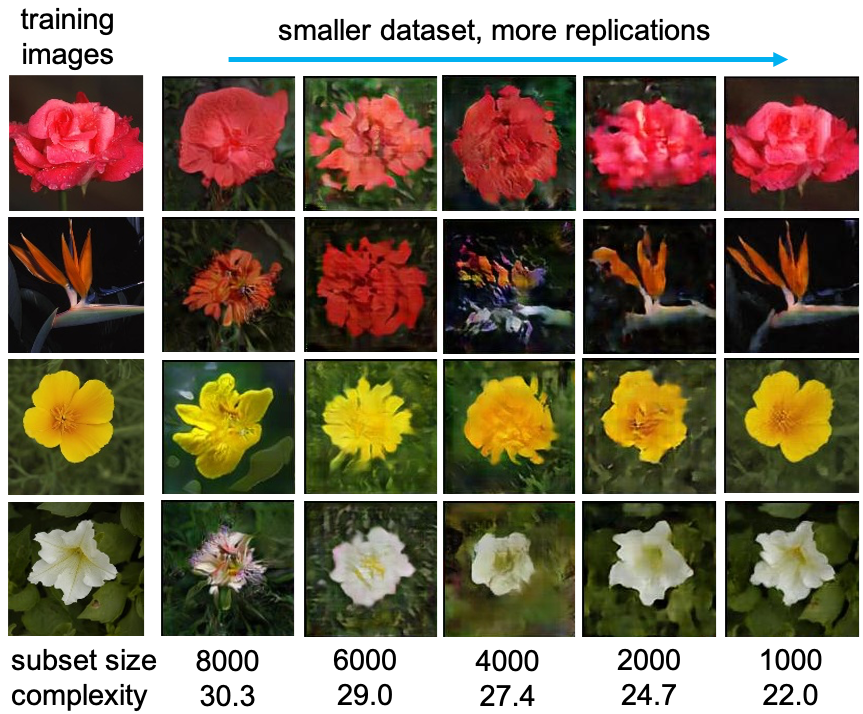}
\end{center}
\vspace{-1em}
\caption{An example of GAN replications in Flower dataset. The complexity is measured by Intrinsic Dimensionality. We study the condition of GAN replication with a particular focus on the effect dataset size/complexity. For a given image synthesis task, when the dataset size decreases, the replication become more prominent.}
\label{fig:teaser-small}
\end{figure}

Generative Adversarial Networks (GANs) has attracted consistent attention since its first proposal in \cite{goodfellow2014generative}. Since then, the photorealism of the synthetic images has seen dramatic improvement with methods like BigGAN \cite{brock2018large}, StyleGAN \cite{karras2019style,karras2020analyzing}, etc. This low cost generation of photo-realistic samples brings new possibility to applications like content creation and dataset augmentation, all of which are based on the assumption that the generator does not merely replicate training data. The GAN replication (or memorization, overfitting \cite{corneanu2019does,corneanu2020computing}) problem, besides its obvious theoretical value, is also practically important. 

A number of recent studies have explored the option of using GANs to augment training datasets to improve the performance of downstream machine learning algorithms \cite{douzas2018effective,wang2019generative,tran2021data,frid2018gan}, especially for medical applications where patients data is scarce. When these data augmentation GANs overfit to the training data, the augmented samples will provide little or no additional value to the downstream algorithms, defying the purpose of augmenting the data. Replicating training samples is also problematic for content creation due to potential copyright infringement \cite{gillotte2019copyright}. This is more problematic when the GAN based face swapping technique is used for preserving patient privacy (de-identification) as used in \cite{zhu2020deepfakes}. If a replication happens during the swapping process, the technique is then protecting one's privacy by costing another's portrait right and potentially spreading mis-information about the individual. Without deep understanding on the mechanism and conditions of GANs replication, one might even found the issue more complicated as the replication itself might not be an intentional action from the user or the creator of the GAN. Thus, it is highly important to further our understanding on the replication behavior of GANs. 

Fortunately, researchers have already started to look into the underlying mechanism and contributing factors of potential GANs replication/memorization \cite{webster2019detecting,radford2015unsupervised,nagarajan2018theoretical,arora2017gans}, which we discuss in Section \ref{sec:related-works}. These works, although providing important insight on the effect of factors like latent code, discriminator complexity, have not yet explored the role of dataset size and complexity in GANs replication. 

In this paper, we attempt to fill this gap by empirically study the relationship between GANs replication and dataset size/complexity. We show that it is possible for GANs to replicate with unmodified training procedure, challenging the common view that GANs tend not to memorize training data under a normal training setup \cite{webster2019detecting}. This finding not only sheds new light on the potential mechanism of GAN replication, but also provides practical guidance on estimating minimal dataset size when new a dataset is being constructed for image synthesis purpose.

\section{Related Works}\label{sec:related-works}

Our study is not the first to study GAN replication. Studies proposing novel GAN architectures (e.g. DCGAN \cite{radford2015unsupervised}, PGAN \cite{karras2017progressive}) usually show that their results are not generated by merely memorizing training samples. Repeated non-memorizing results from novel GANs architectures seem to indicate that with a normal training procedure, GANs memorization is not likely \cite{webster2019detecting}. More specifically, \cite{radford2015unsupervised} validate this information by traversing the latent space and checking for visualization and sharp changes. \cite{webster2019detecting} studies GANs memorization specifically with latent code recovery and conclude that memorization is not detectable for GANs trained without cycle-consistency loss. \cite{nagarajan2018theoretical} provides theoretical context for GANs memorization which further shows that the size of the support of the latent space might lead to unseen latent codes replicating training data. \cite{arora2017gans} studies the relationship between GAN memorization and discriminator size, concluding that the distributions learnt by GANs have significantly less support than real distributions. \cite{yazici2020empirical} then shows that with fixed latent codes during the training process the authors can achieve GAN memorization. 

On the application front, \cite{tinsley2021face} studied identity leakage in the face generation application with GANs. Although the topic is different from ours, the study nevertheless shows significant ethical implication on this issue. Studies like \cite{webster2019detecting,radford2015unsupervised,nagarajan2018theoretical,arora2017gans} aim to address the class imbalance problem by using GANs, due to the random up/down sampling can lead to data replication. However, this assumption of non-replicating behavior of GANs is challenged by our study, leading to more cautious practices when augmenting a small dataset to address the imbalance problem. 

In the previous studies investigating GAN memorization, regardless of active latent code seeking methods, or fixed latent codes, the methods deviate from training procedures originally proposed by each GAN, affecting the external validity of the conclusions. In this study, we achieve GAN replication without modifying training procedures. On the other hand, the previous studies on GAN replication have not yet explore the role of dataset size/compleity. In this study, we reveal a relationship between dataset size, GAN replication and generated image photorealism, providing a more specific guidance on deciding dataset size for image synthesis tasks. 



\section{Problem Definition}

The specific problem concerning this study is the relationship between dataset size/complexity and GAN replication. To formulate the problem mathematically, let us denote a set of training images as $\mathcal{X}$ with each element $\mathbf{X}\in \mathbb{R}^m$. $p_{\mathrm{data}}$ is an empirical data distribution defined on $\mathcal{X}$ (which is usually a uniform discrete distribution over all training images). A generator $G:\mathbb{R}^k\to\mathbb{R}^m$ maps $k$-dimensional latent code random variables $z\sim p_{\mathrm{latent}}$ to synthetic images $G(z)$. Then, the probability that the generator $G$ replicates at level $\alpha$ with respect to metric $d(\cdot,\cdot)$, denoted as $P_{\alpha}(G,d,p_\mathrm{data})$, is,

\begin{equation}\label{eq:replication-definition}
    P_{\alpha}(G,d,\mathcal{X}) = \mathrm{Pr}\left( \left[ \min_{\mathbf{X}\in \mathcal{X}} d(G(z),\mathbf{X}) \right] \leq \alpha \right).
\end{equation}

Ideally, we want to know the exact function $P_{\alpha}(G,d,\mathcal{X})$. But this is impractical as the function defined on a joint domain of all GANs, metrics and datasets is too complex. As described in Section \ref{sec:intro}, we wish to study when does a GAN replicate in terms of dataset size and complexity. Thus, in this study, we limit $G$ and $d(\cdot,\cdot)$ while empirically study the effect of dataset on $P_{\alpha}(G,d,\mathcal{X})$. Let $\mu$ be a set function characterizing the training dataset $\mathcal{X}$, given a generator $G$ and a distance metric $d(\cdot,\cdot)$, the specific function we are to study is, 

\begin{equation}\label{eq:dataset-measure-to-replication}
    f: \mu(\mathcal{X}) \mapsto P_{\alpha}(G,d,\mathcal{X}),
\end{equation}
with a particular focus on $\mu$ being a counting measure and complexity measurement. 

\section{Hypothesis}\label{sec:f-hypothesis}
Although previous studies reported that GAN replication is not observed during the normal training procedures in common datasets, the replication itself should be theoretically possible. Intuitively, when the complexity of a model exceeds data complexity, the model could overfit to the training dataset. 

Let us consider an extreme case of a single image training set. Since there is only one image, as long as the the model is complex enough to capture the within-image complexity, it can fit to the training sample. Since there is only one training image, all the latent codes sampled during the training process are set to map to the same image, achieving the replication defined in Equation (\ref{eq:replication-definition}). When the training dataset becomes larger (populated with non-trivial training samples), the empirical distribution of the dataset grows more complex, making overfitting to the dataset more difficult. However, at this stage, the dataset is not small enough for overfitting, but also has not enough samples to reconstruct the underlying image manifold. One can make analogy to the double maximum frequency threshold for lossless reconstruction of a continuous signal in Nyquist-Shannon Theorem \cite{shannon1949communication}. Once the number of training samples is greater than this effective threshold, it becomes possible to reconstruct the image manifold, given a correct model choice, improving the image quality and reducing replication.

Based on the above analysis, we hypothesize that with dataset size increases, the replication probability $P_{\alpha}(G,d,\mathcal{X})$ decreases. The quality of generated images depicts a U-shape trend with respect to dataset size. The quality is first high when the GAN producing faithful replication of training data. Then both replication and image quality decreases as dataset size increases. Further increasing the dataset size will increase the image quality (photorealism) while the GAN replication remains low.

\section{Definition of Replication}
The choice of image distance metric $d(\cdot,\cdot)$ is important for studying GAN replication as it directly defines the meaning of replication. In this study, we use the Euclidean distance in image space as our metric. 

This choice might seem counter-intuitive since numerous previous studies \cite{radford2015unsupervised,theis2015note} explicitly state to avoid using simple Euclidean distance in nearest neighbour (NN) since it can be easily fooled by simple unperceivable color shift, small translation of the image and even dead pixels. Instead, studies have suggested using semantic spaces (like from InceptionV3 deep feature pretrained on ImageNet) for more effective evaluation and memorization check \cite{xu2018empirical,sajjadi2018assessing,lucic2018gans}. 

However, the aforementioned weaknesses of using the Euclidean distance in pixel space is actually a strength in our study. In previous studies, the aim is to show that the proposed GAN does \textit{not} memorize. If the Euclidean metric is used, then as long as the GAN does not produce exact copies of training samples, one can claim the GAN does not merely memorize, which might be too liberal in most applications. However, in our study, the purpose is to show that GANs \textit{do} memorize. Thus using Euclidean distance means replications have to be exact copies (up to a noise level $\alpha$), which will also be treated as memorization in all the other semantic based metrics.

\section{Dataset Size and Complexity}

We use two set functions $\mu_1,\mu_2$ to characterize training set in our study. The first set function $\mu_1$ is to measure dataset complexity. We use intrinsic dimensionality (ID) for this purpose which can be understood as an estimate of the degree-of-freedom of data manifold in the high dimensional pixel space. We use a maximum likelihood estimator of intrinsic dimensionality \cite{levina2005maximum}, which can be defined as,

\begin{equation}
    \mu_1(\mathcal{X}) = \frac{1}{|\mathcal{X}|(k_1-k_2+1)}\sum_{k=k_1}^{k_2}\sum_{\mathbf{X}\in\mathcal{X}}\hat{m}_k(\mathbf{X})
\end{equation}
\begin{equation}
    \hat{m}_k(\mathbf{X})=\left[ \frac{1}{k-1}\sum_{j=1}^{k-1}\log\frac{T_k(\mathbf{X})}{T_j(\mathbf{X})} \right]^{-1}
\end{equation}
where $T_k(\mathbf{X})$ denotes the Euclidean distance from $\mathbf{X}$ to its $k$-th NN in $\mathcal{X}$. $k_1$ and $k_2$ denote the minimum and maximum $k$ used in k-NN, which affects the locality during ID estimation. 

The reason that we define ID in pixel space rather than in any semantic embedding space learnt by a neural network (as in \cite{gong2019intrinsic}) is to match our choice of $d(\cdot,\cdot)$. 

The second set function is the counting measure for dataset size,
\begin{equation}
    \mu_2(\mathcal{X})=|\mathcal{X}|.
\end{equation}

\section{Selection of GAN Architectures}

Since we are only interested in studying the relationship between GAN replication and dataset size/complexity, any well-established GAN architecture is reasonable for our purpose. We choose to use StyleGAN2 \cite{karras2020analyzing} and BigGAN \cite{brock2018large} for their state-of-the-art performance. 

We follow the original training procedure of BigGAN and StyleGAN2. For BigGAN implementation, we use BigGANdeep architecture with adversarial hinge loss, which corresponds to the implementation with highest performance in \cite{brock2018large}. 
 

For StyleGAN2, we use \textit{config-f} in the original paper, which also provides the highest reported performance. The StyleGAN2 \textit{config-f} uses large networks with regularized adversarial logistic non-saturation loss. 
 
\begin{figure*}[]
\begin{center}
\includegraphics[width=1.0\textwidth]{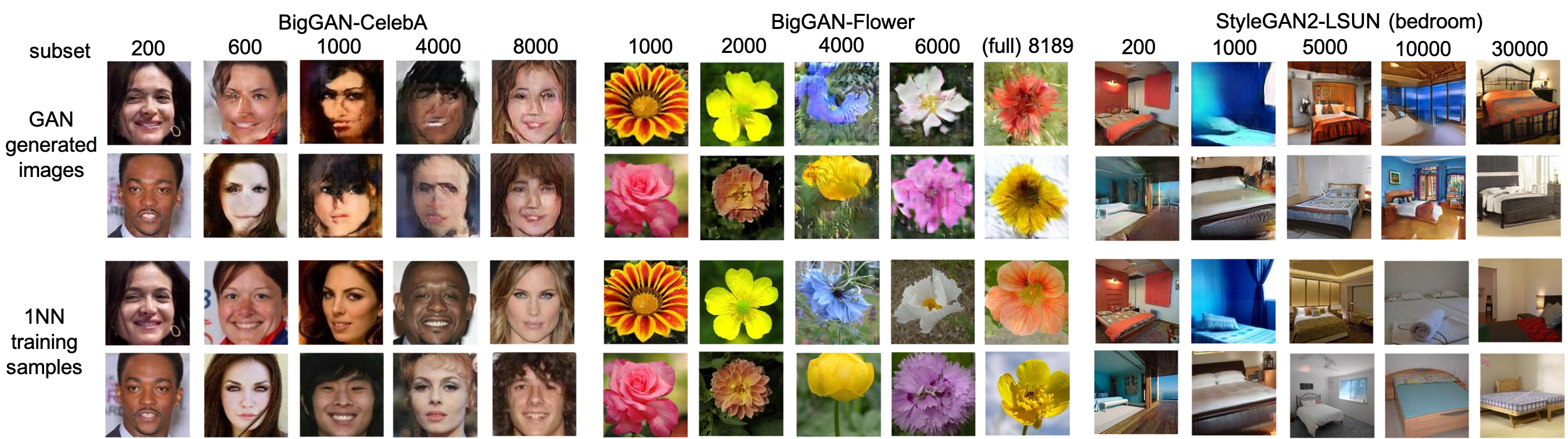}
\end{center}
\vspace{-0.8em}
   \caption{Qualitative results of replication experiments for BigGAN-CelebA, BigGAN-Flower and StyleGAN2-LSUN (bedroom) combination. All images are randomly generated without cherry-picking. Results for all the other experiments are provided in Supplemental Material. For a given GAN and a dataset, at each subset level, a GAN is trained and examined for its replication. This results show that when the dataset size is small, GANs can generate almost exact replication of training data. The replication is gradually alleviated when the dataset size increases.}
\label{fig:sample_replication} 
\end{figure*}

\section{Dataset Setup}\label{sec:dataset}

To estimate the $f$ shown in Equation (\ref{eq:dataset-measure-to-replication}), we need to vary the training sets $\mathcal{X}$. To study the effect of dataset size, we starts with a small subset of the training data and gradually increase the dataset size to observe the change in GANs replication $P_{\alpha}(G,d,\mathcal{X})$. To study the effect of the dataset complexity, we uses multiple datasets for different objects type, combining with various sizes for each set, we build a set of training datasets with different complexities and study their relationship with GAN replication. 

The specific dataset used in our study are CelebA \cite{liu2015faceattributes}, LSUN-bedroom \cite{yu2015lsun}, Oxford Flower 17 \cite{Nilsback08}. This collection of training dataset provides a wide range of variety on the dataset complexity, from simple images like faces to complex scenes like bedrooms. 

From each dataset, we create multiple levels of subsets with different number of samples randomly selected. The specific number of samples depends on the GAN replication trend on the dataset. The sizes of subset levels used in our study are shown in Table \ref{table:dataset-subsize}. 

\begin{table}
\begin{center}
\begin{tabular}{|l|c|}
\hline
Datasets        & Subset size \\
\hline\hline
CelebA          & 200*, 600*, 1,000, 4,000, 8,000 \\
LSUN-bedroom    & 200, 1,000, 50,00, 10,000, 30,000$^\dagger$ \\
Flower          & 1,000, 2,000*, 4,000, 6,000*, 8,189\\
\hline
\end{tabular}
\end{center}
\vspace{-0.8em}
\caption{Subset levels used in our experiments for different datasets. *Subset level only used in BigGAN experiments. $^\dagger$Subset level only used in StyleGAN2 experiments. }
\label{table:dataset-subsize}
\end{table}

\section{Experimental Setup}
To examine the relationship between the dataset size/complexity and GAN replication, we trained BigGAN and StyleGAN2 on each of the subset levels defined in Section \ref{sec:dataset}. 

The images from Flower, CelebA and LSUN-bedroom datasets are first center-cropped and scaled to 128$\times$128 resolution. The RGB channles of training images are z-score normalized. 

To examine the GAN replication, 1,024 samples are first generated for each trained generator. Given a generated sample, we find its NN in the corresponding training set with Euclidean distance in the original pixel space. Then the percentage of generated samples whose NN distance $<\alpha$ is used as the estimate of $P_{\alpha}(G,d,\mathcal{X})$. We report results for $\alpha=8{,}000$ in the main paper,  which is the loosest threshold we found consistent with human perception of replication. Additionally, results for $\alpha=9{,}000$, $10{,}000$ are provided in Supplemental Material to show the effect of different $\alpha$ values on $P_{\alpha}(G,d,\mathcal{X})$. 

For each subset, we calculate the maximum likelihood estimate of its ID \cite{levina2005maximum}\footnote{we use a python implementation \url{https://gist.github.com/mehdidc/8a0bb21a31c43b0cbbdd31d75929b5e4/}.} with $k_1=10$ and $k_2=20$. Before calculating ID, we down-scale the image to 32$\times$32 as it shorten the runtime with no significant difference comparing to the 128$\times$128 version. The empirical evidence supporting this down-scale operation is also provided in the Supplemental Material. 

Both ID and GAN replication calculation require the use of a NN algorithm. We use Faiss \cite{JDH17}\footnote{\url{https://github.com/facebookresearch/faiss}} for its highly efficient implementation of exact NN. 

\begin{figure*}[th!]
\begin{center}
\includegraphics[width=0.95\textwidth]{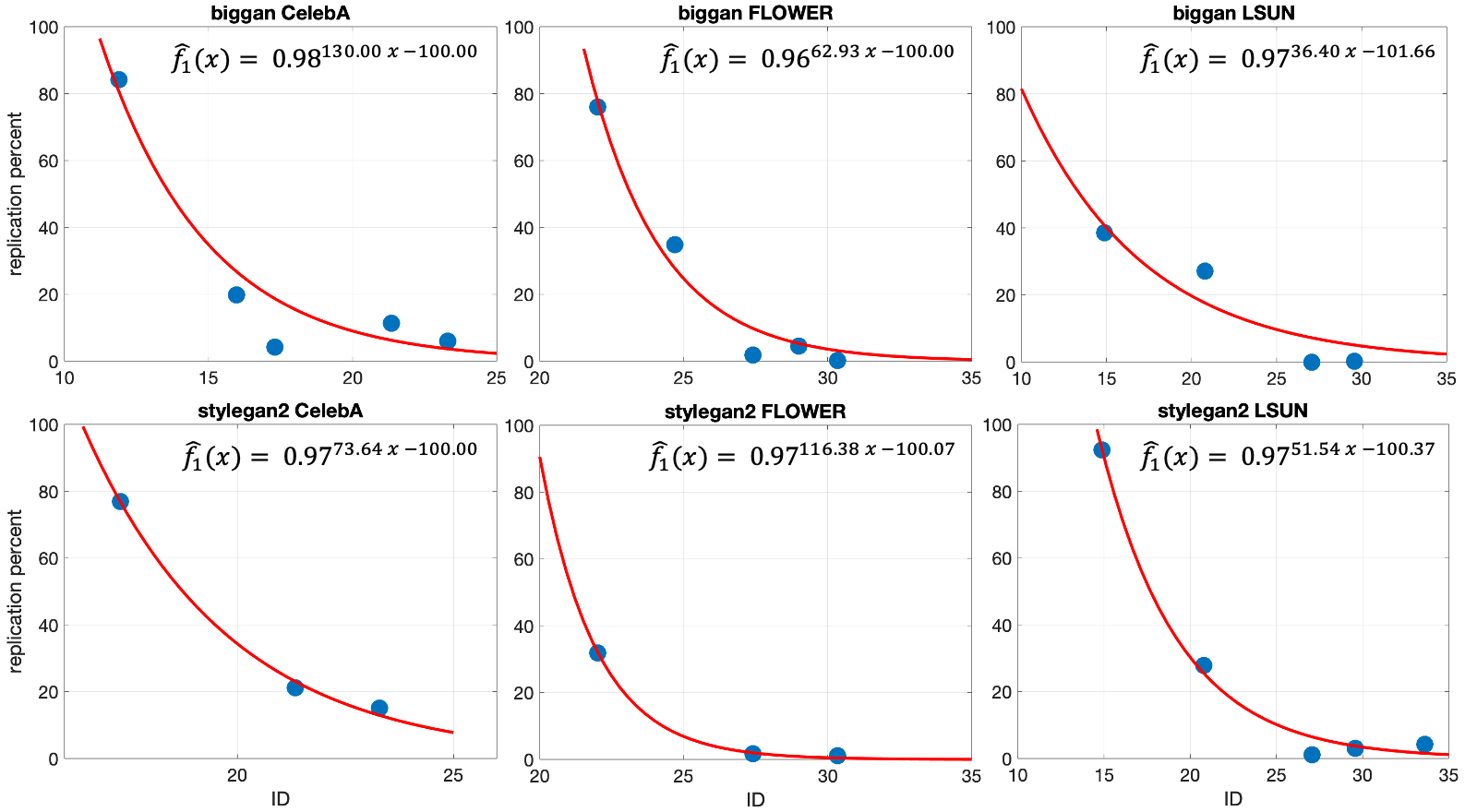}
\end{center}
\vspace{-0.8em}
   \caption{Scatter plots and curve fitting for dataset ID vs GAN replication percentage at each subset level for BigGAN and StyleGAN2 trained on CelebA, Flower and LSUN-bedroom. Regardless of GAN architecture or dataset, the results show a common exponential decay trend. Among the three model parameters to be estimated, the exponential decay factor $a$ and predictor translation $c$ are both shared across GAN architecture and datasets.}
\label{fig:id-vs-replication}
\end{figure*}

\subsection{Measuring perceptual quality}
To measure the perceive quality of the generated images, we run a behavioral experiment with human subjects on Amazon Mechanical Turk. One may wonder the necessity of running a behavioral experiment while the evaluation metric such as Fréchet Inception Distance (FID) is available and well accepted. The reason for not using FID is that it does not necessarily reflects the perceived image quality from human subjects. The supporting evidence for this decision can also be found in the Supplemental Material.

We first randomly sampled 100 images per subset level per GAN per dataset. Each image is rated by 9 subjects for its image quality. A 5-point (Excellent-Good-Fair-Poor-Terrible) Likert scale is used. The rating criteria is described in the Supplemental Material. 

\section{Result and Analysis}

Our hypothesis described in Section \ref{sec:f-hypothesis}, is that when dataset size increases, the dataset complexity increases (before it converges to the underlying complexity of the ground-truth distribution), and the replication percentage decreases. Additionally, we hypothesize that the quality of generated samples first decreases then increases with increasing number of training samples.

\begin{table}[t]
\begin{center}
\begin{tabular}{|l|c|c|c|c|c|}
\hline
GAN         & Datasets  & $R^2_{f_1}$    & $R^2_{g}$ & $R^2_{f_2}$\\
\hline\hline
BigGAN      & Flower    & 0.9739    & 0.9995    & 0.9709\\
StyleGAN2   & Flower    & 0.9994    & 0.9995    & 0.9996\\
BigGAN      & CelebA    & 0.9388    & 0.9999    & 0.8949\\
StyleGAN2   & CelebA    & 0.9965    & 0.9999    & 0.9956\\
BigGAN      & LSUN      & 0.8612    & 1.0000    & 0.8577\\
StyleGAN2   & LSUN      & 0.9930    & 1.0000    & 0.9922\\
\hline
\end{tabular}
\end{center}
\vspace{-0.8em}
\caption{Goodness-of-fit measurement $R^2$ for the ID-replication function $f_1$, intermediate ID-size function $g$ and size-replication function $f_2$ when fitted to our data at all subset levels. The high $R^2$ shows that all the three formulation of $f_1,g$ and $f_2$ are highly effective on modeling the relationship between dataset size/complexity and GAN replication.}
\label{table:r-square-fit-full}
\end{table}

\begin{figure*}[]
\begin{center}
\includegraphics[width=0.95\textwidth]{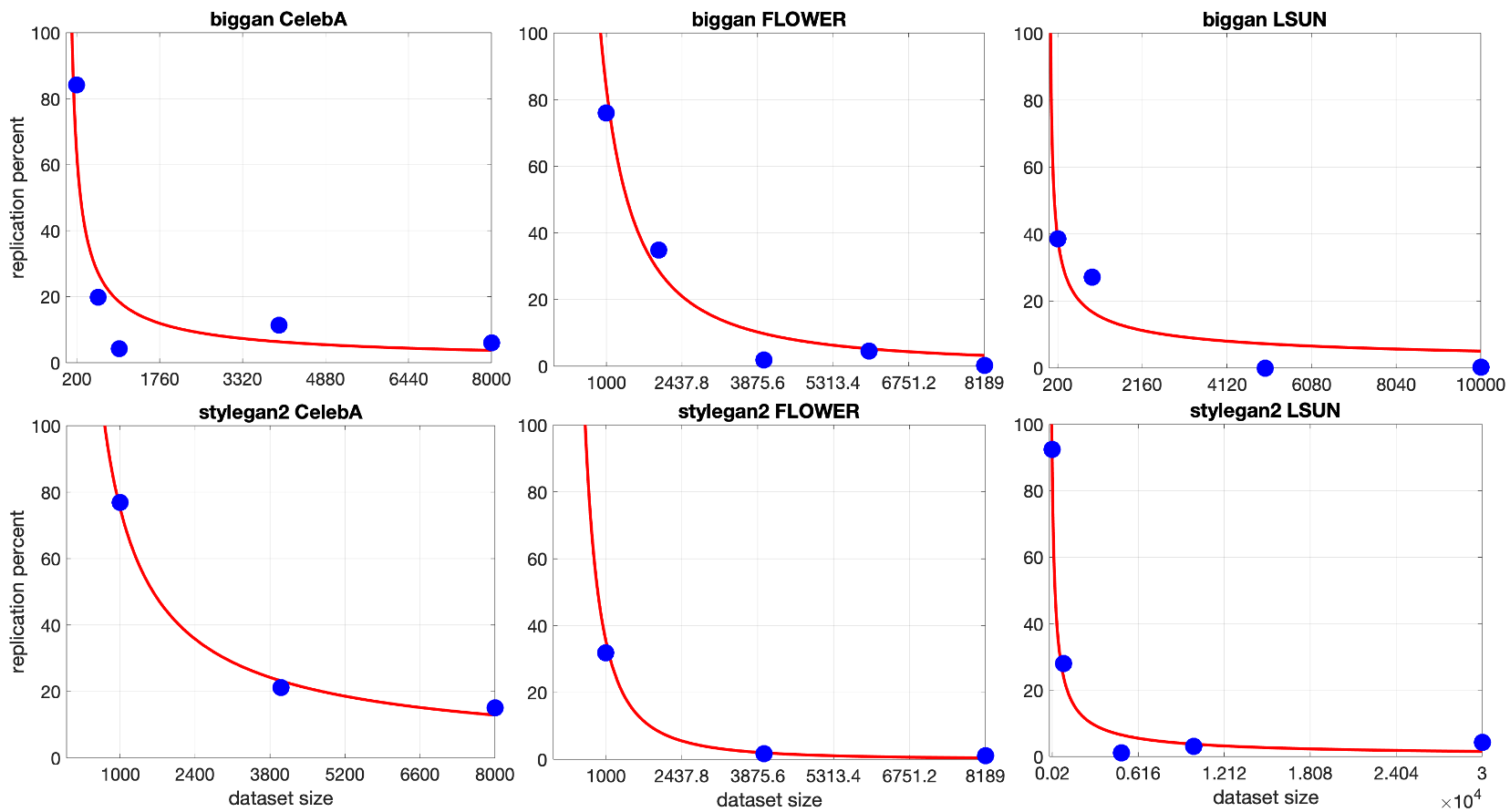}
\end{center}
\vspace{-0.8em}
   \caption{Scatter plots and curve fitting for dataset size vs GAN replication percentage at each subset level. Regardless of GAN architectures or datasets, the results also show a common exponential decay trend similar to the previous ID v.s. replication relationship.}
\label{fig:size-vs-replication}
\end{figure*}

\subsection{Dataset complexity vs. GAN replication}\label{sec:ID-replication}

Figure \ref{fig:id-vs-replication} shows the relationship between dataset complexity and GAN replication percentages for StyleGAN2 and BigGAN trained on CelebA, LSUN-bedroom and Flower datasets. Each point in the figure represents an experiment on a subset level of the corresponding dataset. 

Although the initial replication percentages are different across datasets, the relationship between dataset complexity and GAN replication percentages follows a common trend of exponential decay across all datasets and GAN architectures.

To quantitatively characterize this trend, we fit an exponential function, 

\begin{equation}\label{eq:exp-decay}
    P_{\alpha}(G,d,\mathcal{X}) = f_1(\mu_1(\mathcal{X})) = a^{b \mu_1(\mathcal{X}) - c}
\end{equation}
to each of the dataset experiments (with all the subset levels). Parameters $a$,$b$ and $c$ serves as decay base factor, scaling and translation on the predictor $\mu_1(\mathcal{X})$, respectively. Note that although one may fit the data equally well with $f(x) = a^{x-c}$ with less parameters to estimate, the 3 variable formulation we used here delineates the effect of $a$ and $b$, which will be an important factor to enable one-shot prediction on GAN replication as shown in Section \ref{sec:predict-replication}. 

Figure \ref{fig:id-vs-replication} also shows that the estimated parameters $\hat{a}$, $\hat{b}$ and $\hat{c}$ of Equation (\ref{eq:exp-decay}) fitted to each of GAN-dataset combinations. Despite with different datasets and GAN architectures, the complexity-replication curves share similar exponential decay factor $a$ and predictor translation $c$ (which can also be understood as response scaling). The former falls in to the range of 0.96 to 0.98 and latter at almost exactly 100.0. With only one parameter $b$ to be determined, we can then predict the full curve when there is only one subset level available for the dataset, which will be shown in Section \ref{sec:predict-replication}. 

We also provide measurement of goodness-of-fit ($R^2_{f_1}$) of our $f_1$ formulation when fitted to all the subset levels in Table \ref{table:r-square-fit-full}, which are greater 0.9 in almost all the experiments, supporting the effectiveness of our $f_1$ formulation. 

\begin{figure*}[h!]
\begin{center}
\includegraphics[width=1.0\textwidth]{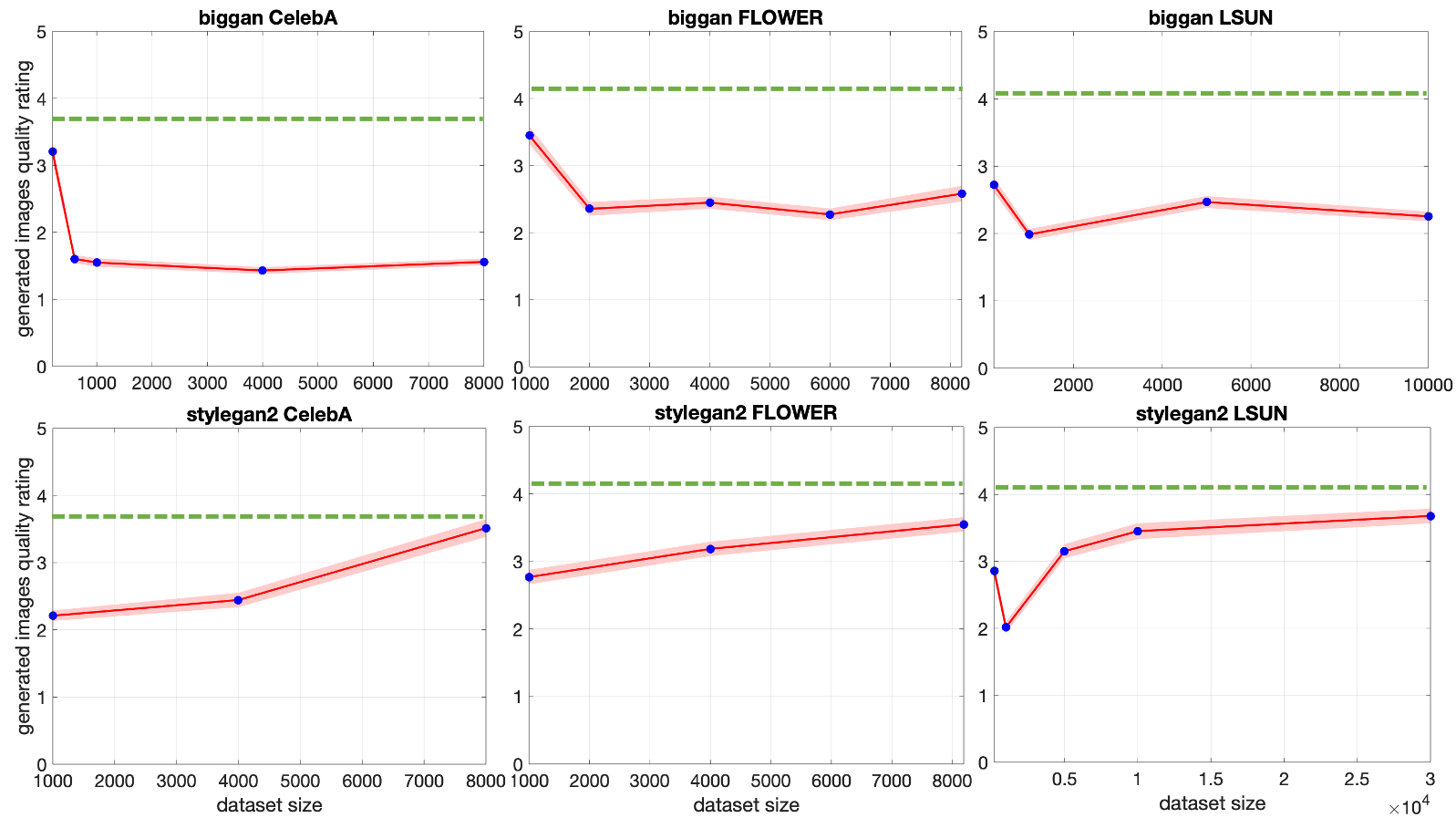}
\end{center}
\vspace{-0.8em}
   \caption{Results of the AMT behavioral experiment testing the relationship between dataset size and perceptual image quality. Average perceptual quality of the image with 95\% confidence interval is provided. Green dashed line indicates the average perceptual quality for real images in the dataset. The image quality is high when the dataset size is small and GANs producing exact replication of the training data, except for the StyleGAN2-CelebA and StyleGAN2-Flower experiments. }
\label{fig:amt-perceptual-quality}
\end{figure*}

\subsection{Dataset size vs. GAN replication}
Figure \ref{fig:size-vs-replication} shows the relationship between dataset size and GAN replication percentages for StyleGAN2 and BigGAN trained on CelebA, LSUN and Flower datasets. Each point in the figure represents an experiment on a subset level of the corresponding dataset.

Since we already have $f_1$ defined in Section \ref{sec:ID-replication} that maps the dataset complexity measurement $\mu_1(\mathcal{X})$ to replication percentage $P_{\alpha}(G,d,\mathcal{X})$, only a change of variable is needed to define the function $f_2$ which maps from dataset size $\mu_2(\mathcal{X})$ to GAN replication percentages. We define an intermediate function $g:\mu_1(\mathcal{X})\mapsto\mu_2(\mathcal{X})$ modeling the relationship between the dataset size and complexity. For $g$, we formulate it a natural exponential function, 

\begin{equation}\label{eq:exp-decay-size-id}
    \mu_2(\mathcal{X}) = g(\mu_1(\mathcal{X})) = \alpha e^{\beta \mu_1(\mathcal{X})}.
\end{equation}

As shown in Table \ref{table:r-square-fit-full} $R^2_g$ column, when fitted to the full subset levels of all three datasets, the model yields $R^2$ close to 1 in all experiments, showing its effectiveness. 


Combining Equation \ref{eq:exp-decay} and inverse of $g$ from Equation \ref{eq:exp-decay-size-id}, we have $f_2$ as,

\begin{equation}\label{eq:exp-decay-size-rep}
\begin{split}
    f_2(\mu_2(\mathcal{X})) & = f_1(g^{-1}(\mu_2(\mathcal{X}))) \\ 
                            & = a^{(b/\beta)\ln{(\mu_2(\mathcal{X})/\alpha)}-c}
\end{split}
\end{equation}
for each of the dataset experiments (with all the subset levels). The solid red curve in Figure \ref{fig:size-vs-replication} shows the the model when fitted to all the subset levels per GAN-dataset experiment, with the goodness-of-fit provided in Table \ref{table:r-square-fit-full} $R^2_{f_2}$ column. Similar to previous results, our formulation is highly accurate with $R^2>0.85$ for all the experiments.

\subsection{Dataset size vs. Perceptual Quality }

Figure \ref{fig:amt-perceptual-quality} shows the relationship between dataset size and the perceptual quality of the images generated by the GAN trained on the dataset. 

Among all combination, the BigGAN-Flower, BigGAN-LSUN and StyleGAN2-LSUN clearly shows the full trend described in our hypothesis which is the perceived image quality is first high when the replication happens, then decrease when the dataset size increases but not large enough, finally increase again when the dataset size becomes larger further. 

On the other hand, BigGAN-CelebA, StyleGAN2-CelebA, StyleGAN2-Flower shows partial trend in our hypothesis, with BigGAN-CelebA shows the first half of the trend and StyleGAN2-CelebA, StyleGAN2-Flower showing the second half. This might because even larger subset level is needed for the former and smaller subset level is needed for the latter.

\begin{table}[]
\begin{center}
\begin{tabular}{|l|c|c|c|c|}
\hline
\multirow{2}{*}{GAN} & \multirow{2}{*}{Datasets}  & \multicolumn{3}{c|}{$R^2_{f_1}$} \\\cline{3-5}
    &           & full & 1-shot & 2-shot \\
\hline\hline
BigGAN & Flower     & 0.9739 & 0.8228 & 0.8230 \\
StyleGAN2 & Flower  & 0.9994 & 1.0000 & 1.0000 \\
BigGAN & CelebA     & 0.9388 & 0.8301 & 0.8174 \\
StyleGAN2 & CelebA  & 0.9965 & 0.9958 & 0.9957 \\
BigGAN & LSUN       & 0.8612 & 0.8830 & 0.8798 \\
StyleGAN2 & LSUN    & 0.9930 & 0.9925 & 0.9925 \\
\hline
\end{tabular}
\end{center}
\vspace{-0.8em}
\caption{Prediction results for complexity-replication function $f_1$ in one-shot ($R^2_{f_1}$ 1-shot) and two-shot ($R^2_{f_1}$ 2-shot) setups. The result for using full subset levels ($R^2_{f_1}$ full) is provided for reference.}
\label{table:complexity-replication-loocv}
\end{table}

\begin{table*}[]
\begin{center}
\begin{tabular}{|l|c|c|c|c|c|c|c|}
\hline
\multirow{2}{*}{GAN}    & \multirow{2}{*}{Datasets}  & \multicolumn{2}{c|}{MAE$_{f_1}$(\%)}    & \multicolumn{2}{c|}{MAE$_{f_2}$(\%)} & \multicolumn{2}{c|}{MAE$_{f_2^{-1}}$(\# of samples)}\\\cline{3-8}
            &           & full      &  1-shot    & full      &  1-shot    &full           &  1-shot   \\
\hline\hline
BigGAN      & Flower    & 2.8855    & 1.8424    & 6.2709    & 4.4886    & 6.0206e2      & 2.1008e3 \\
StyleGAN2   & Flower    & 0.2144    & 0.9725    & 0.6265    & 1.0143    & 2.2244e2      & 1.0546e3 \\
BigGAN      & CelebA    & 5.1180    & 12.0495   & 7.4701    & 12.1064   & 2.1400e3      & 2.1982e4 \\
StyleGAN2   & CelebA    & 1.8955    & 1.2442    & 2.0183    & 2.1006    & 4.5370e2      & 7.4419e2 \\
BigGAN      & LSUN      & 6.0261    & 7.3125    & 6.0358    & 7.6807    & 2.4715e6      & 1.4775e7 \\
StyleGAN2   & LSUN      & 2.4826    & 3.0008    & 4.2719    & 4.6232    & 3.0231e3      & 8.6447e2 \\
\hline
\multicolumn{2}{|c|}{median} & 2.6840    & 2.4215    & 5.1538    & 4.5558    & 1.3710e3      &  1.5777e3 \\
\hline
\end{tabular}
\end{center}
\vspace{-0.8em}
\caption{Median Absolute Errors (MAE) of predicting GAN replication from dataset ID (MAE$_{f_1}$), dataset size (MAE$_{f_2}$) and predicting dataset size from replication (MAE$_{f_2^{-1}}$).}
\label{table:mae-loocv}
\end{table*}

\section{One-shot Prediction on GAN Replication}\label{sec:predict-replication}

As described in Section \ref{sec:intro}, a practical purpose of this study is to provide guidance on the dataest size when a new dataset for image synthesis is under construction. This can be achieved by predicting the replication percentage with dataset size/complexity or vice-versa. Ideally, we wish to predict the curve as early as possible (\textit{i.e.} smallest subset level possible) since the dataset is typically collected by adding more images over time. 

\subsection{From dataset ID to replication}

Section \ref{sec:ID-replication} shows that the exponential decay factor $a$ and predictor translation $c$ are shared across experiments, leaving only one parameter $b$ to be estimated. Thus it is possible to estimate $f_1$ for an unseen dataset with as low as one subset level, by using the shared $a$ and $c$. 

To test the ability to predict replication curve for unseen dataset and GAN architectures, we perform an leave-one-out cross-validation (LOOCV). For each GAN-dataset combination shown in Figure \ref{fig:id-vs-replication}, we hold out one combination for testing, using the rest to estimate the shared parameter $a$ and $c$ by averaging $\hat{a}$ and $\hat{c}$ across combination. Then, for the held-out combination, we estimate $b$ using only one smallest subset level (one-shot) or first two smallest levels (two-shot). This procedure simulates the practical application when the current dataset under collection is still at its early stage with very small number of samples. 

Table \ref{table:complexity-replication-loocv} shows the goodness-of-fit measurement of predicted GAN replication curve with one-shot and two-shot setup denoted as $R^2_{f_1}$(1-shot) and $R^2_{f_1}$(2-shot) respectively. For comparison, we also include $R^2_{f_1}$(full) for the model fitted with full subsets levels. The table shows that even predicted from only one subset level, the model fits data very well with all $R^2>0.8$. Comparing to using full subsets, the one-shot prediction only suffers minor performance drop for BigGAN on Flower and CelebA datasets. The two-shot prediction does not improve over one-shot.

Table \ref{table:mae-loocv} MAE$_{f_1}$ shows a more interpretable performance, Mean Absolute Error (MAE) when predicting replication percentage from dataset ID. The results indicate that overall the median error on replication percentage for a given query dataset ID is around 2.42\% with only one subset level needed, which does not deviate significantly from the 2.42\% median error using full subsets, showing the effectiveness of our one-shot prediction method. 

\subsection{From dataset size to replication}

We perform the same LOOCV one-shot test on predicting replication percentage from dataset size using Equation (\ref{eq:exp-decay-size-rep}). Table \ref{table:mae-loocv} MAE$_{f_2}$ shows the results of this experiment. With the median error around 4.56\%, the results show that dataset size is less accurate on predicting replication percentage in our method, which is not surprising due to accumulation of error during the change of variable.

\subsection{From replication to dataset size}

Since Equation (\ref{eq:exp-decay-size-rep}) is invertible, we can also predict dataset size for a given replication percentage. 

\begin{equation}\label{eq:exp-decay3-inv}
    f_2^{-1}(x) = \alpha e^{(\beta/b)[\log_a(x)+c]}
\end{equation}

Table \ref{table:mae-loocv} MAE$_{f_2^{-1}}$ shows the results of this experiment. The mean error is around 1.3K for full subset levels and 1.5K for one-shot prediction, which is relatively small considering the size of modern datasets often range from hundreds of thousands to millions. In this experiment, BigGAN-CelebA and BigGAN-LSUN performs poorly due to large subset level when replication percentage values are close to zero.

\section{Discussion and Conclusion}

In this study, we show that for a given GAN model and synthesis task, GAN replication percentage decays exponentially w.r.t. dataset size and complexity while the image quality depicts an U-shape trend. We also designed a practical tool to predict the number of training sample necessary in one-shot for a given replication percentage or vice-versa, providing guidance on the choice of dataset size for anyone constructing a novel dataset for image synthesis purpose. This discovery of dataset size and replication relationship also deepen our understanding on the underlying mechanism for GAN replication and overfitting. 





{\small
\bibliographystyle{ieee_fullname}
\bibliography{egbib}
}

\end{document}


\title{Supplemental Material
\par
When does GAN replicate? An indication on the choice of dataset size}


\ificcvfinal\thispagestyle{empty}\fi

\twocolumn[{%
\renewcommand\twocolumn[1][]{#1}%
\maketitle
}]

\section{Summary}\label{sec:Summary}

Section \ref{sec:longVerRepFig} provides extended qualitative results on the GANs replication experiments in the main paper. 

Section \ref{sec:multiThresh} provides additional results for the effect of threshold $\alpha$ on the dataset ID-replication curves. 

Section \ref{sec:resolution-comparison} provides additional results for the effect of training image resolution on dataset ID-replication curves. 

Section \ref{sec:ID_differentSizes} shows supporting empirical evidence on our practices to use downscaled images during the calculation of Intrinsic Dimensionality. 

Section \ref{sec:NN_simpCLR_metrics} shows replication results defined in semantic embedding space rather than the native pixel space used in the main paper.

Section \ref{sec:MNIST} provides additional results on MNIST dataset, which shows the limitation of our method when the dataset is extremely simple.

Section \ref{sec:FIRvsBehav} shows results comparison between the FID scores and perceptual image quality from our AMT experiment. 

Section \ref{sec:AMT_criteria} describes our Amazon Mechanical Turk experiment in details.


\section{Extended Qualitative Results for GANs Replication}\label{sec:longVerRepFig}

Figure \ref{fig:sample_replication_biggan_long} and Figure \ref{fig:sample_replication_stylegan2_long} show extended qualitative results of BigGAN and StyleGAN2 replication experiments for CelebA, Flower and LSUN (bedroom) datasets in the main paper. All images are randomly generated without cherry-picking. These results indicate that for a given GAN architecture and dataset, when the dataset size is small, the GAN can generate almost exact replication of training data. The replication is gradually alleviated when the dataset size increases.

\begin{figure}[h]
\begin{center}
\includegraphics[width=1.0\columnwidth]{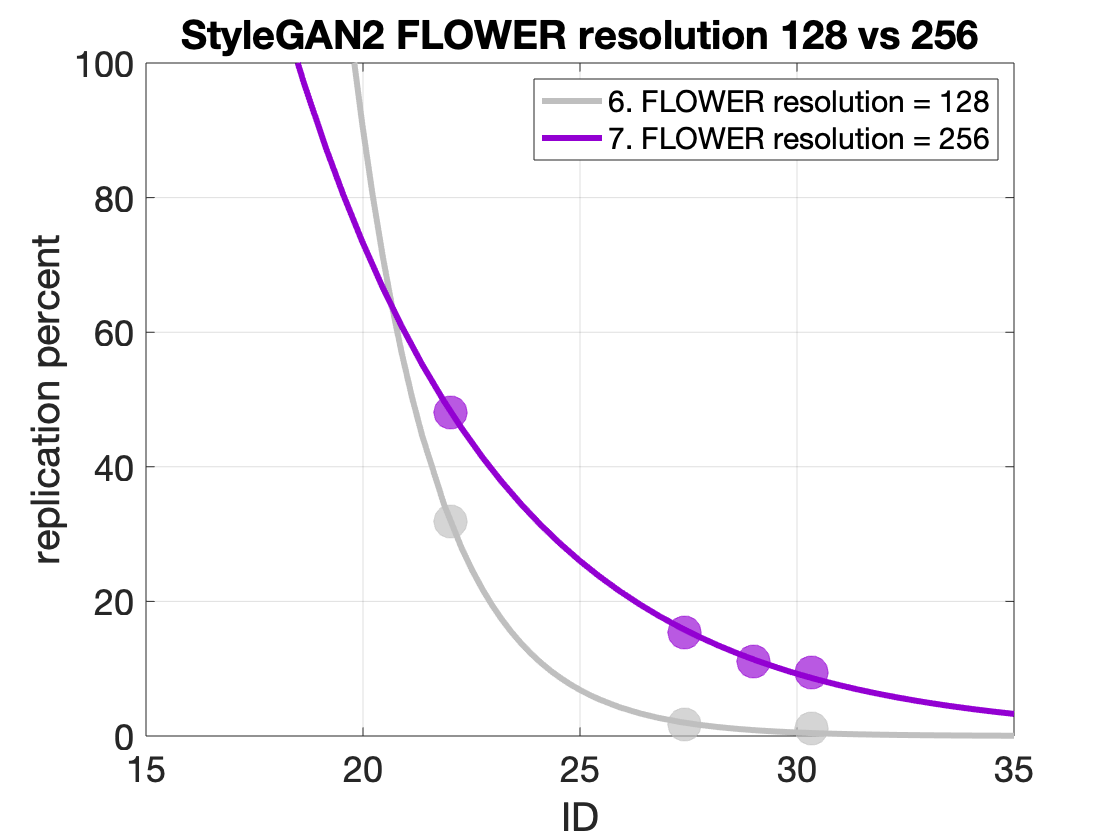}
\end{center}
\vspace{-0.8em}
   \caption{Replication curves for resolutions 128$\times$128 and 256$\times$256 for StyleGAN2 trained on Flower dataset. The trend remains exponential even with different resolutions.}
\label{fig:results-128-vs-256}
\end{figure}

\section{Effects of threshold $\alpha$ on GANs Replication}\label{sec:multiThresh}

In the main paper, we studied the relationship between the dataset size/complexity and GANs replication where the definition of replication is given by 
\begin{equation}\label{eq:replication-definition}
    P_{\alpha}(G,d,\mathcal{X}) = \mathrm{Pr}\left( \left[ \min_{\mathbf{X}\in \mathcal{X}} d(G(z),\mathbf{X}) \right] \leq \alpha \right)
\end{equation}
with threshold $\alpha$ served as the acceptable noise level. Figure 3 of the main paper shows that when $\alpha=8000$, the replication percentage shows a consistent exponential decay trend with respect to the dataset size and complexity, with a shared exponential decay factor. One may now wonder, how does the noise threshold $\alpha$ affect this trend? In this section, we will explore the effect of $\alpha$\footnote{We also use $\alpha$ in the Equation 7, as part of the parameters in the modeling, which is not what we are examining here. This confusion in notation will be addressed in the final version of the main paper.}.

Figure \ref{fig:id-vs-replication-multiThresh} shows the replication-complexity curve same as Figure 3 in the main paper but with threshold at $\alpha=7000,8000,9000,10000$, which shows that decreasing $\alpha$ will leads to a faster decrease of replication percentage. Table \ref{table:multi_thresh_param} shows the parameters $a$, $c$ and $b$ estimated for all the $\alpha$'s. The goodness-of-fit measurement is provided in Table \ref{table:multi_thresh_metric}. 

These results show that trend discovered in the main paper still holds across different thresholds, which means the proposed one-shot prediction will perform similarly well even for different $\alpha$'s. This further implies that for anyone using our one-shot prediction method, the deciding factor of $\alpha$ should be purely driven by the strictness of the replication with minimal concerns on the performance of the one-shot predictor.  




\begin{table}[t]
\begin{center}
\begin{tabular}{|l|c|c|c|c|c|}
\hline
GAN         & Datasets  & $\alpha$  & $\hat{a}$     & $\hat{b}$     & $\hat{c}$  \\
\hline\hline
BigGAN      & Flower    & 7000      & 0.96          & 92.36         & 100.00     \\
StyleGAN2   & Flower    & 7000      & 0.98          & 248.35        & 100.00     \\
BigGAN      & CelebA    & 7000      & 0.99          & 500.00        & 100.00     \\
StyleGAN2   & CelebA    & 7000      & 0.99          & 491.60        & 100.00     \\
BigGAN      & LSUN      & 7000      & 0.98          & 78.71         & 100.00     \\
StyleGAN2   & LSUN      & 7000      & 0.98          & 95.25         & 100.00     \\
\hline
BigGAN      & Flower    & 8000      & 0.96          & 62.93         & 100.00     \\
StyleGAN2   & Flower    & 8000      & 0.97          & 116.38        & 100.08     \\
BigGAN      & CelebA    & 8000      & 0.98          & 130.00        & 100.00     \\
StyleGAN2   & CelebA    & 8000      & 0.97          & 73.65         & 100.00     \\
BigGAN      & LSUN      & 8000      & 0.97          & 36.40         & 101.66     \\
StyleGAN2   & LSUN      & 8000      & 0.97          & 51.55         & 100.37     \\
\hline
BigGAN      & Flower    & 9000      & 0.96          & 43.19         & 100.00     \\
StyleGAN2   & Flower    & 9000      & 0.97          & 61.30         & 100.62     \\
BigGAN      & CelebA    & 9000      & 0.98          & 69.72         & 100.30     \\
StyleGAN2   & CelebA    & 9000      & 0.97          & 32.27         & 101.81     \\
BigGAN      & LSUN      & 9000      & 0.97          & 25.11         & 101.26     \\
StyleGAN2   & LSUN      & 9000      & 0.97          & 31.00         & 101.08     \\
\hline
BigGAN      & Flower    & 10000     & 0.96          & 30.52         & 100.15     \\
StyleGAN2   & Flower    & 10000     & 0.96          & 38.94         & 100.52     \\
BigGAN      & CelebA    & 10000     & 0.97          & 21.65         & 100.71     \\
StyleGAN2   & CelebA    & 10000     & 0.96          & 15.21         & 100.20     \\
BigGAN      & LSUN      & 10000     & 0.96          & 20.46         & 100.61     \\
StyleGAN2   & LSUN      & 10000     & 0.96          & 17.70         & 100.70     \\
\hline
\end{tabular}
\end{center}
\vspace{-0.8em}
\caption{Estimated parameters of the exponential decay model for different thresholds $\alpha$. Under different threshold $\alpha$, parameter $a$,$b$ and $c$ estimated from exponential relationship between dataset ID and GAN replication percentages for StyleGAN2 and BigGAN trained on CelebA, LSUN-bedroom and Flower datasets. For each $\alpha$ value, despite with different datasets and GAN architectures, the complexity-replication curves share similar exponential decay factor $a$ and predictor translation $c$. For 
a given dataset-GAN combination, $b$ decreases as $\alpha$ increases.}
\label{table:multi_thresh_param}
\end{table}

\begin{table}[t]
\begin{center}
\begin{tabular}{|l|c|c|c|c|}
\hline
GAN         & Datasets & $\alpha$   & $R^2$     & MAE       \\
\hline\hline
BigGAN      & Flower   & 7000      & 0.9942     & 0.9029    \\
StyleGAN2   & Flower   & 7000      & 0.9999     & 0.0308    \\
BigGAN      & CelebA   & 7000      & 0.9704     & 2.6241    \\
StyleGAN2   & CelebA   & 7000      & 0.9970     & 1.2866    \\
BigGAN      & LSUN     & 7000      & 0.9606     & 1.7540    \\
StyleGAN2   & LSUN     & 7000      & 0.9992     & 0.5416    \\
\hline
BigGAN      & Flower   & 8000      & 0.9739     & 2.8855    \\
StyleGAN2   & Flower   & 8000      & 0.9994     & 0.2144    \\
BigGAN      & CelebA   & 8000      & 0.9388     & 5.1180    \\
StyleGAN2   & CelebA   & 8000      & 0.9965     & 1.8955    \\
BigGAN      & LSUN     & 8000      & 0.8612     & 6.0261    \\
StyleGAN2   & LSUN     & 8000      & 0.9930     & 2.4826    \\
\hline
BigGAN      & Flower   & 9000      & 0.9250     & 8.2542    \\
StyleGAN2   & Flower   & 9000      & 0.9990     & 6.2406    \\
BigGAN      & CelebA   & 9000      & 0.8638     & 4.5048    \\
StyleGAN2   & CelebA   & 9000      & 0.9993     & 17.2335   \\
BigGAN      & LSUN     & 9000      & 0.7828     & 13.3370   \\
StyleGAN2   & LSUN     & 9000      & 0.9733     & 6.5316    \\
\hline
BigGAN      & Flower   & 10000     & 0.8600     & 12.0084   \\
StyleGAN2   & Flower   & 10000     & 0.9969     & 1.4222    \\
BigGAN      & CelebA   & 10000     & 0.6800     & 11.6200   \\
StyleGAN2   & CelebA   & 10000     & 0.9980     & 0.8923    \\
BigGAN      & LSUN     & 10000     & 0.7911     & 14.2998   \\
StyleGAN2   & LSUN     & 10000     & 0.9339     & 7.9402    \\
\hline
\end{tabular}
\end{center}
\vspace{-0.8em}
\caption{$R^2$ and MAE of the models across different thresholds $\alpha$. $R^2$: goodness-of-fit measurement. MAE: median absolute errors. The model at each row corresponds to the ones in Table \ref{table:multi_thresh_param}. Although there is a minor decrease of $R^2$ with the increase of $\alpha$, the overall $R^2$ values are still close to one, showing that the model is highly effective on approximating the relationship between dataset complexity and GAN replication, independent from the threshold $\alpha$. 
}
\label{table:multi_thresh_metric}
\end{table}

\section{Effect of dataset resolution on GANs replication}\label{sec:resolution-comparison}
One may also wonder the relationship between the image resolution to the replication curve, as we only showed results of 128$\times$128 in the main paper.  Figure \ref{fig:results-128-vs-256} shows the replication results for StyleGAN2 trained on Flower dataset in 128$\times$128 v.s. 256$\times$256 resolutions. As shown in the figure, the exponential trend remains when we increase the resolution to 256.

\section{Dataset ID under Different Image Sizes}\label{sec:ID_differentSizes}
To calculate Intrinsic Dimensionality (ID) of the images, we downscale images from 128$\times$128 to 32$\times$32 to save computational resources. In this section, we provide supporting experiments on comparing the dataset ID between these two resolutions across all the datasets used in the main paper.  

Table \ref{table:ID-differentSizes} provides the results of this comparison, which shows that the dataset ID for low-resolution images does not differ significantly from their high-resolution counterparts, albeit systematically lower as expected, which means that the observed trend of exponential decay shown in the main paper with 32$\times$32 will not change once the ID is calculated with high-resolution samples. 

\begin{table}[t]
\begin{center}
\begin{tabular}{|l|c|c|c|c|c|}
\hline
\multirow{2}{*}{Datasets} & \multirow{2}{*}{Subset size}  & \multicolumn{2}{c|}{ID} \\\cline{3-4}
    &           & $32\times32$ & $128\times128$ \\

\hline\hline
Flower   & 1000         & 22.02            & 22.66  \\
Flower   & 2000         & 24.70            & 25.77  \\
Flower   & 4000         & 27.41            & 28.30  \\
Flower   & 6000         & 28.99            & 30.30  \\
Flower   & 8189         & 30.34            & 31.49  \\
\hline
CelebA   & 200          & 11.90            & 12.23  \\
CelebA   & 600          & 15.97            & 16.50  \\
CelebA   & 1000         & 17.30            & 18.01  \\
CelebA   & 4000         & 21.34            & 22.23  \\
CelebA   & 8000         & 23.29            & 24.28  \\
\hline
LSUN     & 200          & 14.87            & 15.33  \\
LSUN     & 1000         & 20.80            & 21.64  \\
LSUN     & 5000         & 27.06            & 28.31  \\
LSUN     & 10000        & 29.57            & 30.94  \\
LSUN     & 30000        & 33.60            & 35.26  \\
\hline
\end{tabular}
\end{center}
\vspace{-0.8em}
\caption{Comparison on Intrinsic Dimensionality (ID) calculated with $32\times32$ and $128\times128$ samples, for each dataset and subset levels used in the main paper. }
\label{table:ID-differentSizes}
\end{table}

\section{Replications in Semantic Spaces}\label{sec:NN_simpCLR_metrics}

In Section 5 of the main paper, the replication is defined in RGB pixel space since any replication in the RGB space will imply replication in other commonly used semantic spaces, but not \textit{vice versa}. In this section, to further illustrate this point, we provide qualitative results for replications defined in other semantic spaces, including the InceptionV3 \cite{szegedy2016rethinking} semantic embedding and SimCLR contrastive embedding \cite{chen2020simple}. The former is widely used as for visual semantic representation for natural images and is also part of FID calculation \cite{heusel2017gans}. The latter is shown to be an effective embedding (learnt with self-supervision) for a variety of downstream tasks.

\subsection{InceptionV3 Semantic Space}

Our method to define the InceptionV3 semantic space is the same as in FID calculation. More specifically, for each image, synthetic or real, we pass it to the InceptionV3 network pre-trained on ImageNet after center-croping, resizing and normalization. The 2048 dimensional feature output from the last pooling layer is used as the final embedding vector. The replication of a query (generated) image is then defined as its nearest neighbour in the InceptionV3 embedding space using Euclidean distance whose distance to the query image is smaller than the threshold $\alpha$. 

We provide qualitative comparison of the replication in RGB space to the ones in InceptionV3 semantic space for the Flower dataset (at subset level=1000) in Figure \ref{fig:NN_in_different_space_inceptionv3}. We used $\alpha=15$ for InceptionV3 space which yields a comparable replication percentage as $\alpha=10000$ in RGB space (87.11\% and 89.75\% for BigGAN, and 66.89\% and 63.08\% for StyleGAN2, in Inception and RGB space respectively). The replications found in RGB space gives nearly perfect matching (thus a perceptual replication), while features in InceptionV3 space can only capture part of the image feature and thus leads to unsatisfactory matching results.


\subsection{SimCLR Contrastive Space}

Features in SimCLR space are acquired by passing images into SimCLR network. The network is trained on each dataset used in our experiments without pre-training. Besides real images, synthetic images are also used during the training to improve its embedding quality.
We define the positive pair as two transformed versions of a single image. The transformations we allowed are Gaussian blur and small affine transformations \footnote{For gaussian blur, we use kernel widith=51. For affine transformations, we use \textit{RandomAffine} in torchvision package with degrees=10, translate=None, scale=None, shear=10. This yields an image perceptually similar to the original sample.}. The negative pair is defined by any two different images in a mini-batch. We use the embedding after the encoder before the projection head as our feature. Each feature is of 2048 dimensions.

We provide qualitative results of the replication in RGB space to the ones in SimCLR semantic space for the FLOWER dataset (at subset level=1000) in Figure \ref{fig:NN_in_different_space_simclr}. We used $\alpha=0.35$ for SimCLR space which yields a comparable replication percentage as $\alpha=10000$ in RGB space (91.99\% and 89.75\% for BigGAN, and 61.42\% and 63.08\% for StyleGAN2, in SimCLR and RGB space respectively). Similar to the results in the Inception space, the replications in SimCLR space can only capture part of the image characteristics and thus cannot be qualified as perceptual replications.

\subsection{Image2StyleGAN space}
We also tested using the combination of pixel and semantic space metrics described in Image2StyleGAN paper \cite{abdal2019image2stylegan} with proper weight and normalization. As shown in Figure \ref{fig:image2stylegan-space-results}, the replication results with Image2StyleGAN space is almost identical to the one obtained with the RGB-metric. 

\begin{figure}[h]
\begin{center}
\includegraphics[width=1.0\columnwidth]{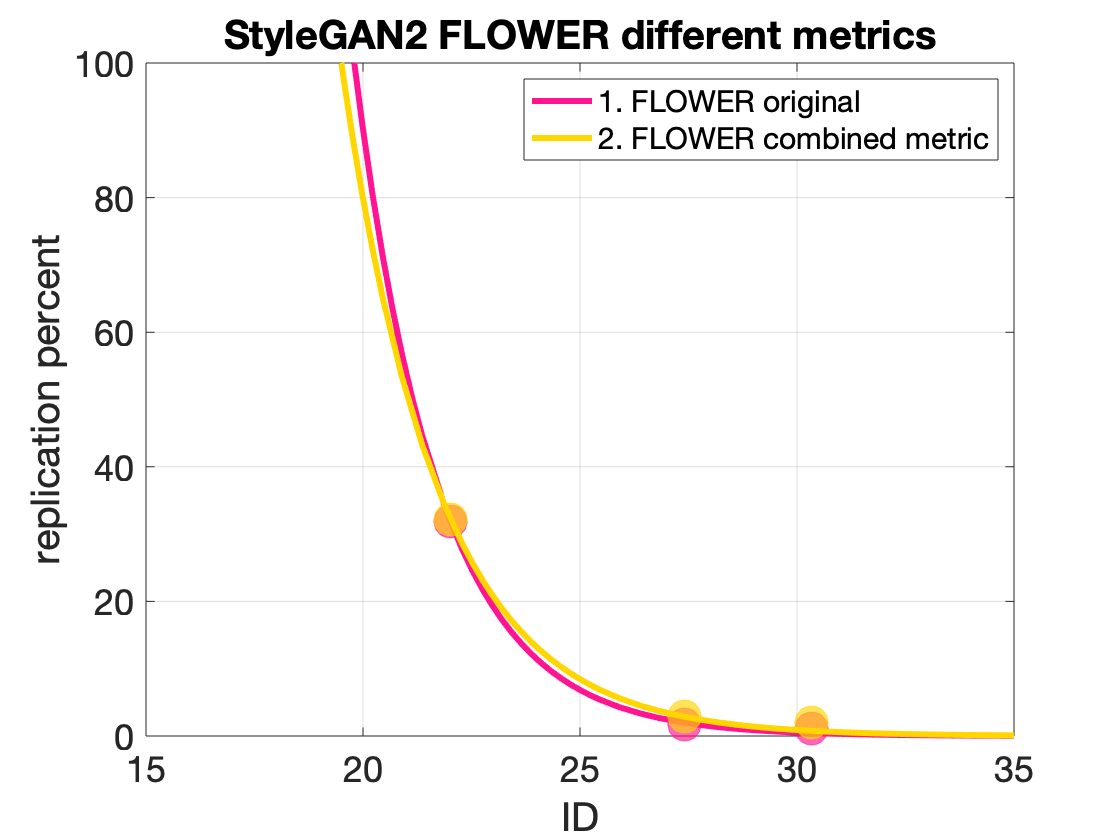}
\end{center}
\vspace{-0.8em}
   \caption{Replication curve when the distance metric is defined in Image2StyleGAN space (combined metric) v.s. RGB space (original).}
\label{fig:image2stylegan-space-results}
\end{figure}

\section{Additional Results for MNIST dataset}\label{sec:MNIST}
We also conducted additional experiment on MNIST dataset with the same setup as described in the main paper. As shown in the Figure, both BigGAN and StyleGAN2 can produce exact replication even with the full dataset (the curve is approximately flat). This is due to the simplicity of the dataset (with ID=12.7 at full which is close ID=12.23 for CelebA-200). This prompts caution on applying our method for extremely simple dataset.

\begin{figure}[h]
\begin{center}
\includegraphics[width=1.0\columnwidth]{figure/mnist-result.png}
\end{center}
\vspace{-0.8em}
   \caption{GAN replication curves for BigGAN and StyleGAN2 trained on MNIST comparing to other datasets.}
\label{fig:image2stylegan-space-results}
\end{figure}


\section{FID v.s. Perceptual Image Quality}\label{sec:FIRvsBehav}

To show the necessity of using Behavioral Experiment results instead of Fréchet Inception Distance (FID)  as perceived image quality metric, Figure \ref{fig:humanCmpFID-quality} compares the Amazon Mechanical Turk (AMT) rating with FID for StyleGAN2 and BigGAN on CelebA, LSUN-bedroom and Flower datasets under each subset level.

\section{Details for Amazon Mechanical Turk Experiment}\label{sec:AMT_criteria}

In the main paper, we describe that the perceptual image quality is measured with a behavioral experiment on Amazon Mechanical Turk. The purpose of this experiment is to see how the perceived quality of the generated images changes with respect to subset levels. 

For each generator trained at each subset level of a GAN-dataset combinations, we randomly generate 100 images. For each dataset, we also randomly select 100 real images for references. 

Each of the real and synthesized images are rated by 9 AMTurkers. To ensure the quality of the rating, we limited the Workers to have HIT approval rate at least 95\% and minimal 500 HIT's approved. Each HIT task contains total of 5 different images. A Worker needs to rate all 5 images to proceed to the next one. Maximum 10 minutes is allowed to complete each task. All the workers are compensated 0.05 USD (with 0.01 fee to AMT) for each task they completed. 

We use 5-point Likert scale to measure the perception of image quality from human subjects. For each image, the following instruction is given, 

\begin{displayquote}
    \textit{Read the task carefully and inspect the image. Choose the appropriate level of quality that best suits the image:}
    \begin{enumerate}
        \item \textit{\textbf{[Excellent]} the image looks no different than a real image.}
        \item \textit{\textbf{[Good]} the image looks close to a real image but there are some issues barely noticeable.}
        \item \textit{\textbf{[Fair]} the image has some small issues but overall looks close to a real image.}
        \item \textit{\textbf{[Poor]} the image has obvious issues but I can see what is the image about.}
        \item \textit{\textbf{[Terrible]} the image looks terrible and I cannot see what it is about.}
    \end{enumerate}
\end{displayquote}

The Workers can use mouse and keyboard to select the level most suitable for the image. The average completion time for each task (5 images) is 1 minute 39 seconds. 

To process the data, we first use integer coding to convert the quality level to numbers, with \textit{Excellent} at 5 and \textit{Terrible} at 1. To report the result, we aggregate the answers from the Workers by calculating the mean and its 95\% confidence interval for each subset level for all the GAN-dataset combinations, which is shown in Figure 5 of the main paper.

\begin{figure*}[]
\begin{center}
\includegraphics[width=1.0\textwidth]{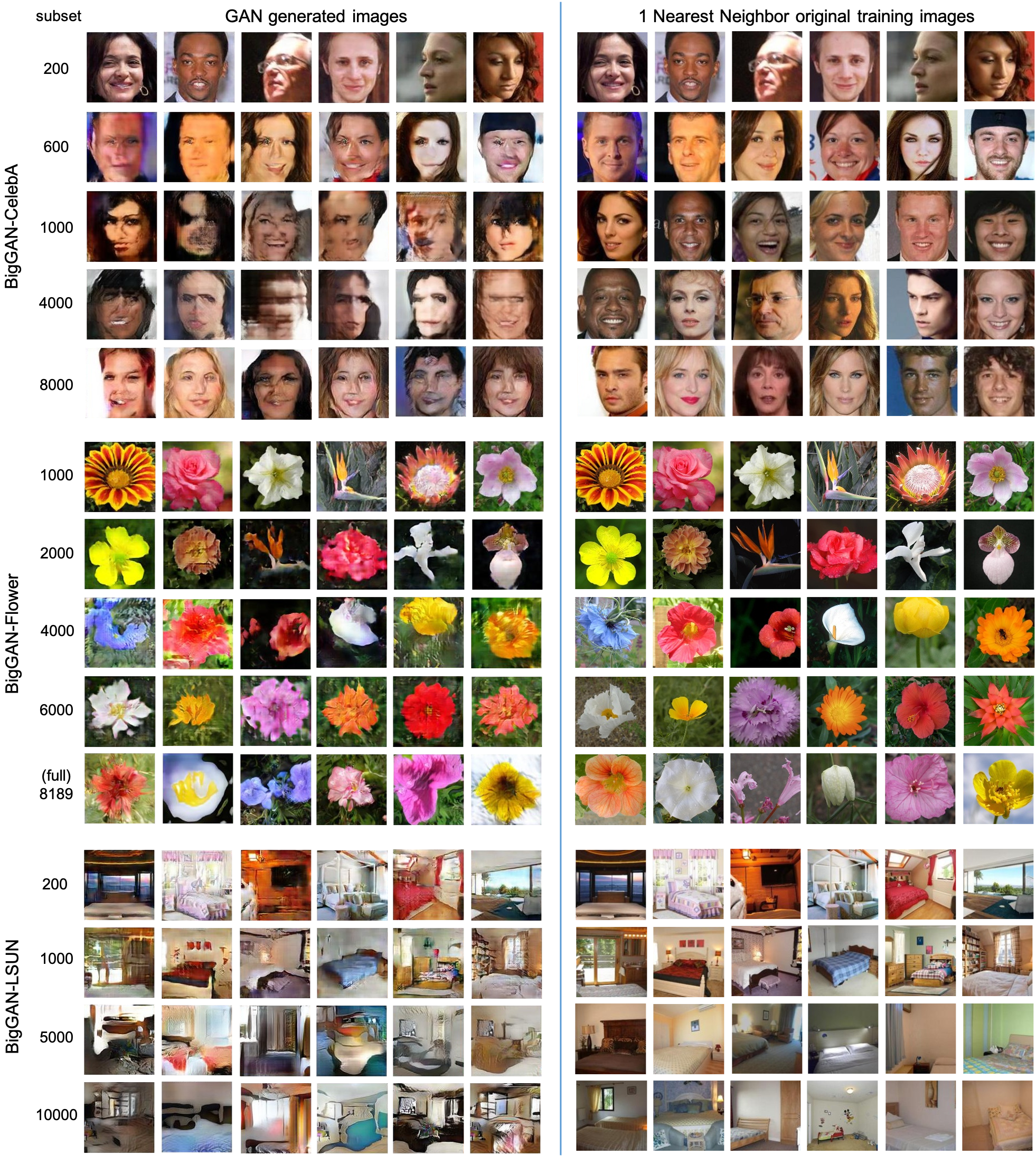}
\end{center}
\vspace{-0.8em}
   \caption{Qualitative results of replication experiments for BigGAN-CelebA, BigGAN-Flower and BigGAN-LSUN (bedroom) combination. All images are randomly generated without cherry-picking. For the BigGAN and a given dataset, at each subset level, a BigGAN model is trained and examined for its replication. This results show that when the dataset size is small, BigGANs can generate almost exact replication of training data. The replication is gradually alleviated when the dataset size increases.}
\label{fig:sample_replication_biggan_long}
\end{figure*}

\begin{figure*}[]
\begin{center}
\includegraphics[width=1.0\textwidth]{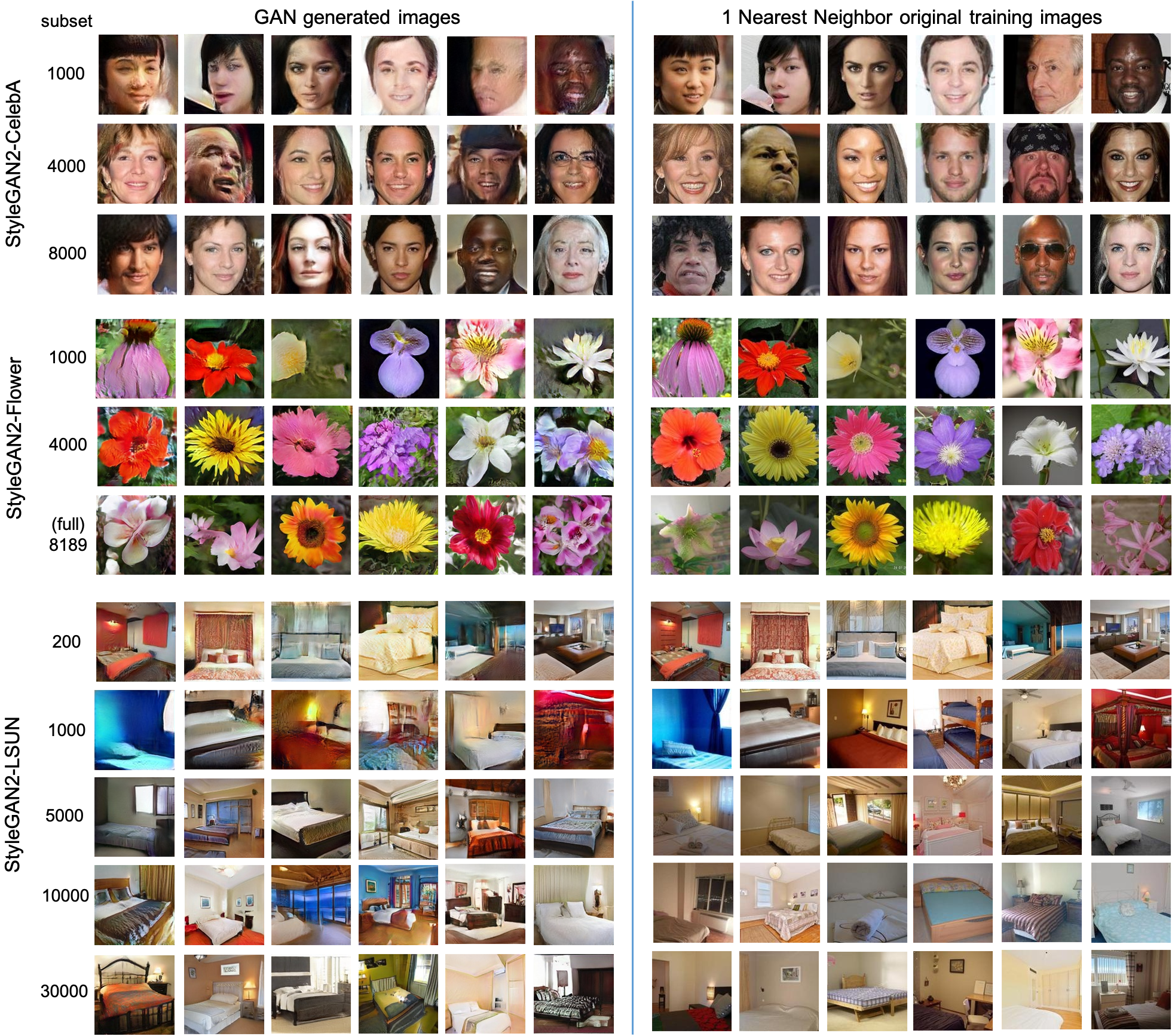}
\end{center}
\vspace{-0.8em}
   \caption{Qualitative results of replication experiments for StyleGAN2-CelebA, StyleGAN2-Flower and StyleGAN2-LSUN (bedroom) combination. All images are randomly generated without cherry-picking. For the StyleGAN2 and a given dataset, at each subset level, a StyleGAN2 model is trained and examined for its replication. This results show that when the dataset size is small, StyleGAN2 can generate almost exact replication of training data. The replication is gradually alleviated when the dataset size increases.}
\label{fig:sample_replication_stylegan2_long}
\end{figure*}

\begin{figure*}[th!]
\begin{center}
\includegraphics[width=0.95\textwidth]{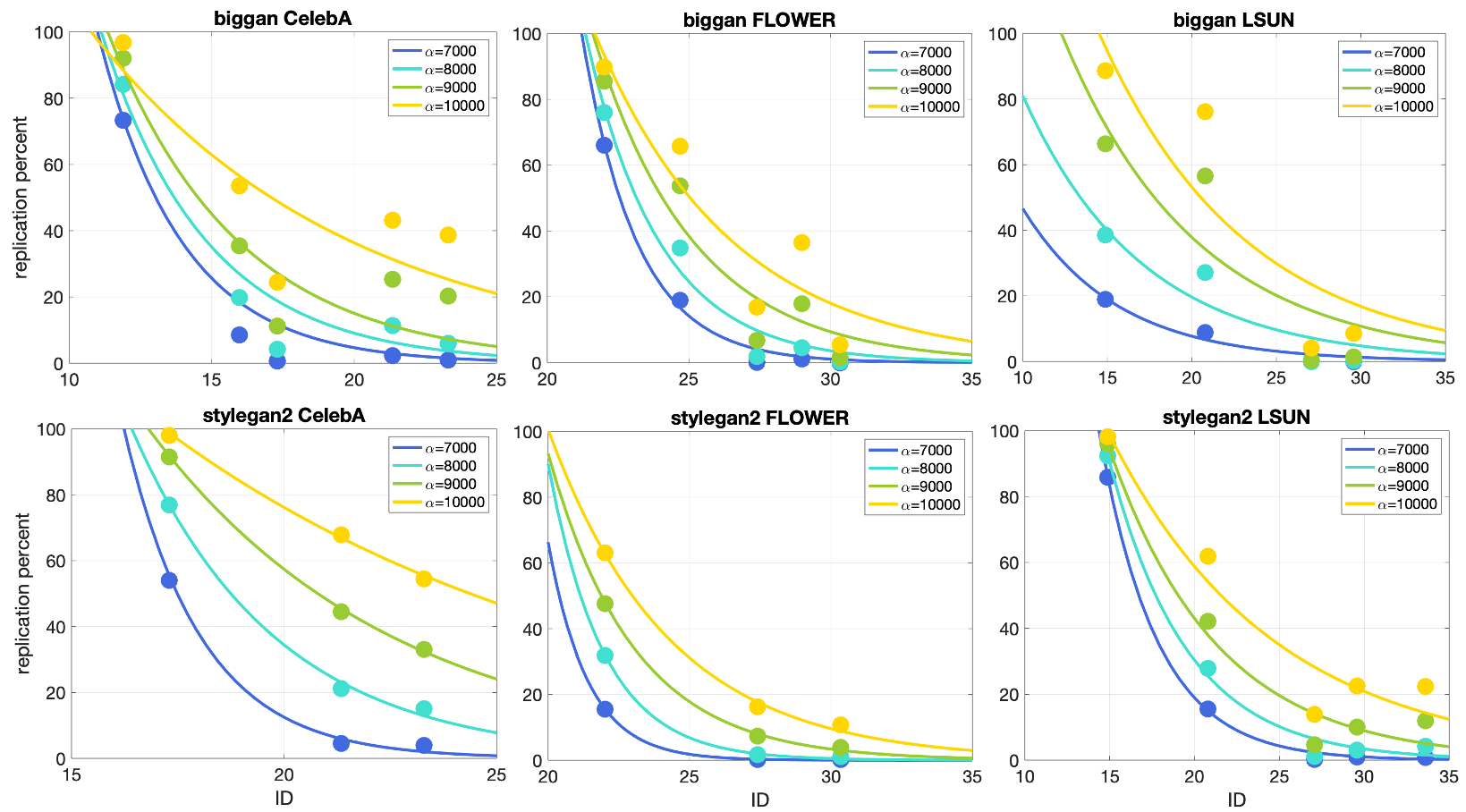}
\end{center}
\vspace{-0.8em}
   \caption{Scatter plots and curve fitting for dataset ID vs GAN replication percentage at each subset level for BigGAN and StyleGAN2 trained on CelebA, Flower and LSUN-bedroom, with different thresholds $\alpha$. For a given $\alpha$, regardless of GAN architectures, datasets or $\alpha$ values, the results show a common exponential decay trend. For each plot with same dataset and GAN architecture but increasing threshold values $\alpha$, the fitted curves gets farther away from the origin, indicating a decreasing scaling $b$ value.}
\label{fig:id-vs-replication-multiThresh}
\end{figure*}

\begin{figure*}[h!]
\begin{center}
\includegraphics[width=1.0\textwidth]{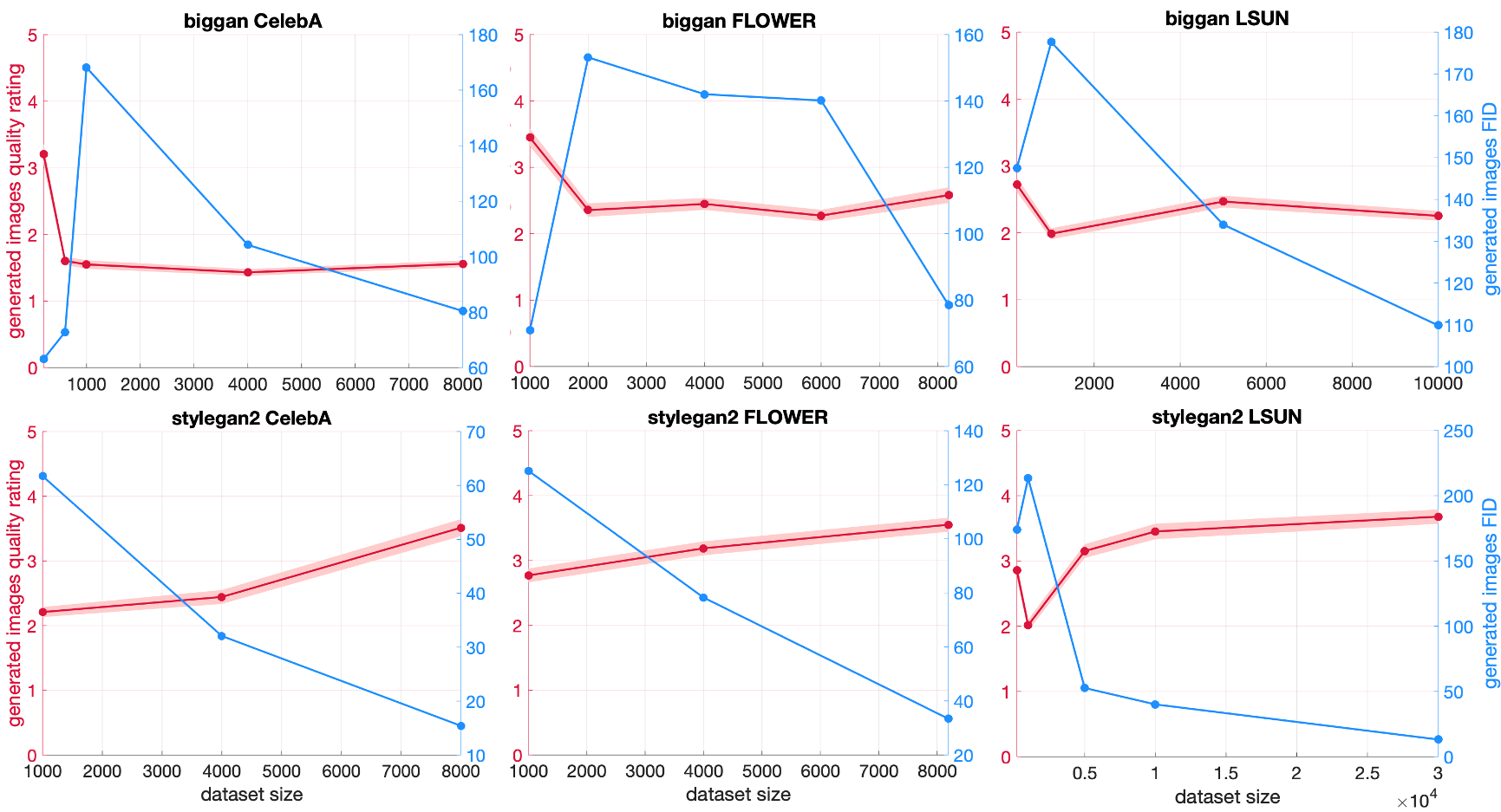}
\end{center}
\vspace{-0.8em}
   \caption{Comparison between perceptual image quality and FID. Each figure shows the curves of perceptual image quality (red) and FID (blue) with respect to the subset levels for each GAN-dataset combination. Although in most experiments, the FID correlates (anticorrelates) with the perceptual image quality acquired from AMT behavioral experiments, the two shows different trends in BigGAN-CelebA experiment, where the image quality improves significantly from subset level 1000 to 8000 according to FID, but only slightly for perceptual quality rating. Note that for FID, the smaller the better and for perceptual image quality, the higher the better. 
   }
\label{fig:humanCmpFID-quality}
\end{figure*}

\begin{figure*}[h!]
\begin{center}
\includegraphics[width=1.0\textwidth]{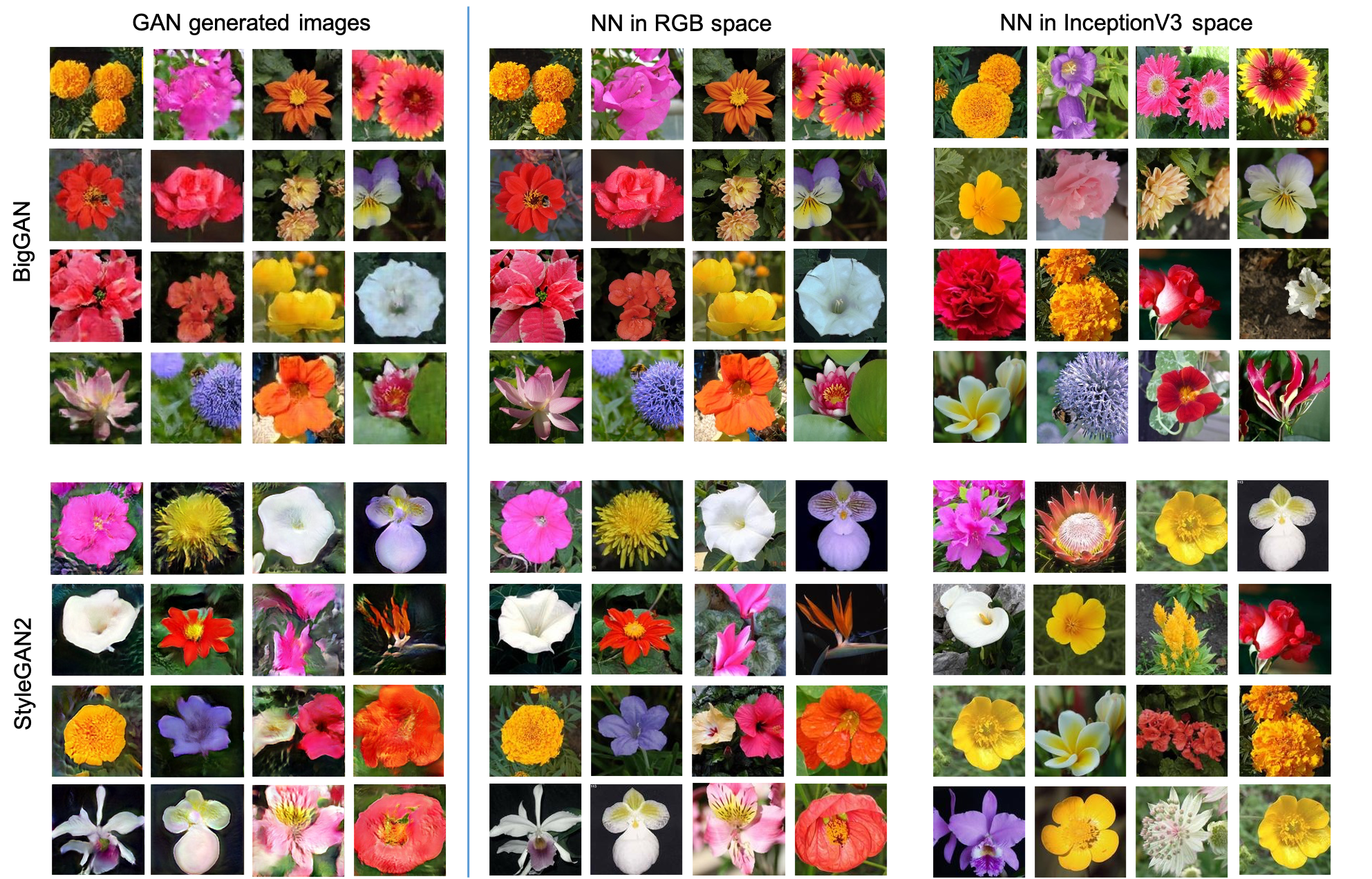}
\end{center}
\vspace{-0.8em}
   \caption{Comparison on image replications in RGB and InceptionV3 space. FLOWER dataset with subset 1000 images are used. RGB space gives nearly perfect matching, while InceptionV3 space can only capture parts of the image feature and thus leads to matching results inconsistent with human perceptions. When graded by a single human rater, $59.86\%$ of the InceptionV3 replications for BigGAN and $94.01\%$ replications for StyleGAN2 are unsatisfactory.}
\label{fig:NN_in_different_space_inceptionv3}
\end{figure*}

\begin{figure*}[h!]
\begin{center}
\includegraphics[width=1.0\textwidth]{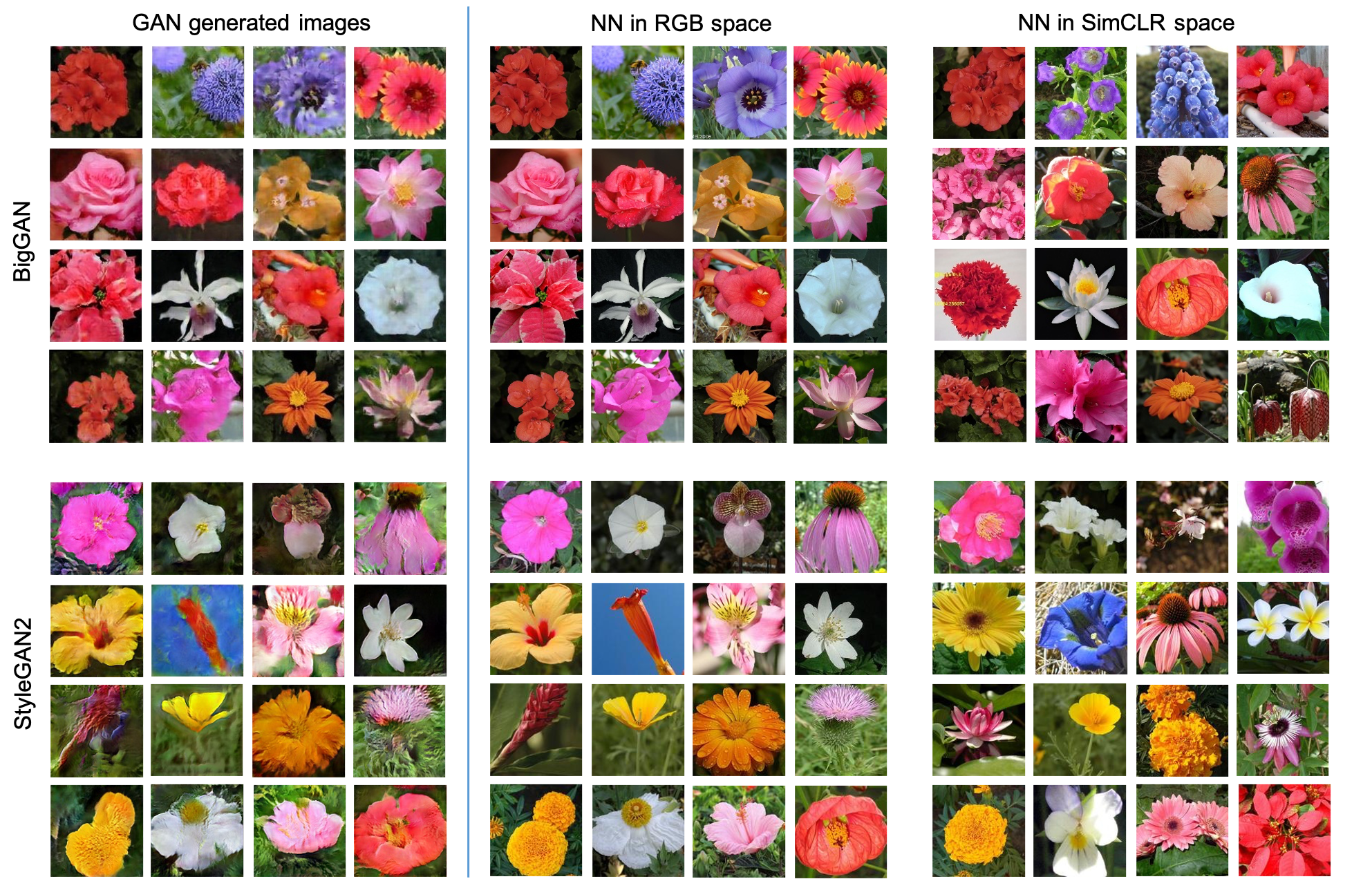}
\end{center}
\vspace{-0.8em}
   \caption{Comparison on image replications in RGB and SimCLR space. FLOWER dataset with subset 1000 images are used. RGB space gives nearly perfect matching, while SimCLR space can only capture parts of the image feature and thus leads to matching results inconsistent with human perceptions. When graded by a single human rater, $10.51\%$ of the SimCLR replications for BigGAN and $39.43\%$ for StyleGAN2 are unsatisfactory.}
\label{fig:NN_in_different_space_simclr}
\end{figure*}

{\small
\bibliographystyle{ieee_fullname}
\bibliography{egbib}
}